\documentclass[sigconf, nonacm]{acmart}
\AtBeginDocument{%
  }

\setcopyright{acmlicensed}

\acmDOI{10.1145/nnnnnnn.nnnnnnn} 
\acmISBN{978-x-xxxx-xxxx-x/YY/MM} 
\acmConference[GECCO '25]{The Genetic and Evolutionary Computation Conference 2025}{July 14--18, 2025}{Málaga, Spain}
\acmYear{2025}
\copyrightyear{2025}

\usepackage[ruled,linesnumbered]{algorithm2e}
\usepackage{multirow}
\usepackage{subfigure}
\usepackage{colortbl}
\begin{document}

\title{Surrogate Learning in Meta-Black-Box Optimization: \\ A Preliminary Study}

\author{Zeyuan Ma}
\email{scut.crazynicolas@gmail.com}
\orcid{0000-0001-6216-9379}
\affiliation{%
  \institution{South China University of Technology}
  \city{Guangzhou}
  \state{Guangdong}
  \country{China}
}

\author{Zhiyang Huang}
\email{scut.hzy@gmail.com}
\orcid{0009-0002-6499-0237}
\affiliation{%
  \institution{South China University of Technology}
  \city{Guangzhou}
  \state{Guangdong}
  \country{China}
}
\author{Jiacheng Chen}
\email{jackchan9345@gmail.com}
\orcid{0000-0002-7539-6156}
\affiliation{%
  \institution{South China University of Technology}
  \city{Guangzhou}
  \state{Guangdong}
  \country{China}
}
\author{Zhiguang Cao}
\email{zhiguangcao@outlook.com}
\orcid{0000-0002-4499-759X}
\affiliation{%
  \institution{Singapore Management University}
  \city{Singapore}
  \country{Singapore}
}
\author{Yue-Jiao Gong}
\email{gongyuejiao@gmail.com}
\authornote{Corresponding Author}
\orcid{0000-0002-5648-1160}
\affiliation{%
  \institution{South China University of Technology}
  \city{Guangzhou}
  \state{Guangdong}
  \country{China}
}
\renewcommand{\shortauthors}{Ma et al.}

\begin{abstract}
Recent Meta-Black-Box Optimization (MetaBBO) approaches have shown possibility of enhancing the optimization performance through learning meta-level policies to dynamically configure low-level optimizers. However, existing MetaBBO approaches potentially consume massive function evaluations to train their meta-level policies. Inspired by the recent trend of using surrogate models for cost-friendly evaluation of expensive optimization problems, in this paper, we propose a novel MetaBBO framework which combines surrogate learning process and reinforcement learning-aided Differential Evolution algorithm, namely Surr-RLDE, to address the intensive function evaluation in MetaBBO. Surr-RLDE comprises two learning stages: surrogate learning and policy learning. In surrogate learning, we train a Kolmogorov-Arnold Networks (KAN) with a novel relative-order-aware loss to accurately approximate the objective functions of the problem instances used for subsequent policy learning. In policy learning, we employ reinforcement learning (RL) to dynamically configure the mutation operator in DE. The learned surrogate model is integrated into the training of the RL-based policy to substitute for the original objective function, which effectively reduces consumed evaluations during policy learning. Extensive benchmark results demonstrate that Surr-RLDE not only shows competitive performance to recent baselines, but also shows compelling generalization for higher-dimensional problems. Further ablation studies underscore the effectiveness of each technical components in Surr-RLDE. We open-source Surr-RLDE at https://github.com/GMC-DRL/Surr-RLDE.     
\end{abstract}




\maketitle

\section{Introduction}
Black-Box Optimization (BBO)~\cite{bbosurvey} holds a wide range of applications and requires effective optimization techniques such as Evolutionary Computation~(EC) to address them.
Recently, Meta-Black-Box Optimization (MetaBBO)~\cite{metabbo_survey_ours, metabbo_survey_others} has gained incremental research interests by showing a feasible paradigm of combining learning-based techniques~(e.g., reinforcement learning~\cite{rl_survey}) and EC to reduce expertise dependent tuning of human-crafted EC methods when addressing BBO problems. MetaBBO comprises bi-level optimization~\cite{ma2024metabox}: a meta-level policy interacts with a low-level BBO process by configuring the low-level EC method for optimization and collecting feedback signal from the performance improvement during the low-level optimization process. The meta-level policy can then be meta-learned by maximizing the accumulated feedback signals over a problem distribution to attain a generalizable and flexible policy capable of optimizing novel BBO problems, with minimal expertise requirement.   

Although MetaBBO methods~\cite{les, rldas, lde,gleet,guo2024configx} have shown promising performance in enhancing traditional EC methods, their training processes unavoidably consume massive computational resources. This attributes to the inherent workflow of existing MetaBBO methods, where the optimization trajectory is repeatedly sampled through the interaction of meta-level policy and low-level optimization which involves numbers of function evaluations on the training optimization problems. Considering the development of surrogate models in recent data-driven evolutionary algorithms~\cite{ddea_survey}, the core research question of this paper comes out: can surrogate modeling techniques be combined with MetaBBO approach to reduce the massive number of function evaluations consumed during the meta learning process?

To explore this research question, in this paper, we propose the use of surrogate models as substitutes for the original low-level function evaluation of MetaBBO. Specifically, we adopt Kolmogorov-Arnold Network (KAN)~\cite{liu2024kan} as the surrogate model for its validated robustness in formula representation~\cite{kan_comparison}, and train each training problem instance a separate KAN model. To improve the fitting accuracy during the surrogate learning, we additionally design a relative-order-aware loss function to ensure that surrogate model maintains the relative objective value ranks between different sample points.    
As a preliminary study, we integrated the learned KAN-based surrogate model into a MetaBBO framework to evaluate its performance. In our setup, the MetaBBO framework is designed to learn a reinforcement learning-based meta-level policy capable of dynamically configuring Differential Evolution (DE~\cite{de_original}), namely Surr-RLDE. Specifically, we construct a comprehensive configuration space including both the operator selection and the parameter tuning for DE mutation operator. The meta-level policy is meta-trained to provide flexible mutation configurations for the low-level DE hence maximizing the optimization performance sorely dependent on KAN-based surrogate models of the training problem instances. Extensive benchmark results demonstrate that incorporating surrogate models with meta-learning achieves comparable or even superior optimization performance to existing MetaBBO baselines, while saving all function evaluations needed during training. Additional ablation studies further validate that design choices in Surr-RLDE contribute to the satisfactory performance indeed, e.g., the proposed novel Relative-Order-Aware(ROA) loss function and the selection of a KAN as the surrogate model.

\section{Related Works}

\subsection{Meta-Black-Box Optimization}\label{sec:2.1}
Meta-Black-Box Optimization (MetaBBO)~\cite{metabbo_survey_ours, metabbo_survey_others} aims to reduce the manual effort typically required for adaptive tuning in traditional optimization algorithms. Although there is a variety of MetaBBO approaches that adopt different learning techniques in the meta-level, in this paper, we focus on a particular kind of MetaBBO approach, those training their policies with RL. Specifically, MetaBBO can be modeled as a Markov Decision Process (MDP), $\mathcal{M} := \left< \mathcal{S}, \mathcal{A}, \mathcal{T}, \mathcal{R} \right>$. At each time step $t$, the meta-level RL-Agent $\pi_\omega$ takes current optimization state $s^t \in \mathcal{S}$ as input and accordingly outputs action $a^t = \pi_\omega(s^t) \in \mathcal{A}$. In MetaBBO, $a^t$ is applied to low-level strategy $\lambda$ to construct a completed optimizer, which is to interact with optimization tasks directly. After $a^t$ is executed, next state $s^{t+1}$ is return according to transition dynamic $\mathcal{T}$ and reward $r^t$ can be calculated.  Given a dataset of problem instances $\mathbb{P}$, the meta-level objective of $\pi_\omega$ is defined to maximize the expectation of $r^t$ over $\mathbb{P}$. Mathematically:
\begin{equation}
    \pi_\omega^* = \underset{\pi_\omega}{argmax} \left[\mathbb{E}_{f \in \mathbb{P}}\left[\sum_{t=1}^T r^t \right]\right]
\end{equation}
where $r^t$ quantifies the performance gain at each time step, which usually demonstrates the relative performance gain between consecutive steps, which is highly dependent on the relative fitness among candidate solutions.

Among existing MetaBBO methods that utilize RL as the meta-level learning algorithm, the action space varies widely, including operator selection (e.g., DEDDQN~\cite{deddqn}, DEDQN~\cite{dedqn}), algorithm selection (e.g., RLDAS~\cite{rldas}), hyperparameter tuning (e.g., LDE~\cite{lde}, GLEET~\cite{gleet}), and even algorithm generation (e.g., SYMBOL~\cite{chen2024symbol}). In addition to RL-based approaches, other methods have also been investigated. For instance, LES~\cite{les} and LGA~\cite{lga} adopt CMA-ES~\cite{cma_original} as the meta-level strategy, while B2Opt~\cite{li2023b2opt} and POM~\cite{pom} employ supervised learning for meta-level training.

Despite these advances, current MetaBBO methods are predominantly applied to synthetic optimization tasks~\cite{ma2024metabox}, such as CEC benchmark functions~\cite{cec_problem}, particularly during the training phase. This limitation arises from the high computational cost of MetaBBO training, which often requires a large number of additional fitness evaluations. For example, the authors of SYMBOL reported the need for an extra 528M evaluations during training, rendering it impractical for many real-world applications where evaluation is costly. To address this issue, incorporating surrogate models as substitutes for fitness evaluations during training could significantly expand the applicability of MetaBBO to more complex and expensive scenarios.
\vspace{-0.5em}
\subsection{Surrogate Model Learning}
Surrogate models are usually used as simplified approximations of more complex and computationally expensive functions~\cite{surrogate_survey1, surrogate_survey2}. The motivation for using surrogate models comes from the prohibitive expense of direct evaluations in many real-world scenarios, such as engineering design optimization~\cite{surrogate_edo}, scientific simulations~\cite{surrogate_ss}, and hyperparameter tuning for machine learning algorithms~\cite{surrogate_hpo}.
Training surrogate models typically involves collecting a set of samples from the original function, consisting of input-output pairs, and using these samples to fit the surrogate model. This process often employs regression techniques to minimize the prediction error in the training data. Common approaches to surrogate model learning include Gaussian Processes (GP)~\cite{gp} and various deep-learning-based methods such as neural networks~\cite{nn_surrogate}. Each of these models has unique strengths: GPs excel in providing uncertainty estimates, making them particularly useful for small-scale optimization tasks, while neural networks are favored for their scalability and ability to handle high-dimensional inputs in large datasets.

Surrogate models have demonstrated success in diverse applications. For instance, in design optimization~\cite{surrogate_do}, surrogate models are used to approximate expensive finite element simulations, enabling rapid prototyping and testing. In evolutionary algorithms~\cite{ddea_survey}, surrogate models are used to replace the expensive fitness evaluation function or serve as an approximation when only historical data are available. In our method, we utilized KAN to build surrogate models, which are employed to replace the real fitness evaluation during the training phase of MetaBBO.

\subsection{Kolmogorov-Arnold Network}
\paragraph{Formulation} KAN (Kolmogorov-Arnold Network)~\cite{liu2024kan} is a type of neural network inspired by the Kolmogorov-Arnold representation theorem. This theorem states that any multivariate continuous function can be expressed as the composition of a finite number of continuous univariate functions. Specifically, the representation is given by:
\begin{equation}
    f(\mathbf{x})=f\left(x_1, \cdots, x_n\right)=\sum_{q=1}^{2 n+1} \Phi_q\left(\sum_{p=1}^n \phi_{q, p}\left(x_p\right)\right)
\end{equation}
where $\Phi$ and $\phi$ are univariate functions. When applied in a multilayer fashion, the formula for a KAN becomes:

\begin{equation}
    \text{KAN}(x)=\Phi^L\left(\Phi^{L-1}\left(\ldots\left(\Phi^2\left(\Phi^1(x)\right)\right)\right)\right)
\end{equation}
where $x$ denotes the input vector and $\Phi_k$ represents the transformation matrix of the $k$-th layer. For example, the calculation in layer 1 is:

\begin{equation}
    \Phi^1(x) = \left[\begin{array}{cccc}\phi_{11} & \phi_{12} & \cdots & \phi_{1 n} \\ \phi_{21} & \phi_{22} & \cdots & \phi_{2 n} \\ \vdots & \vdots & \ddots & \vdots \\ \phi_{m 1} & \phi_{m 2} & \cdots & \phi_{m n}\end{array}\right] \cdot\left[\begin{array}{c}x_1 \\ x_2 \\ \vdots \\ x_n\end{array}\right]
\end{equation}

To make $\phi_{ij}$ trainable and easy to implement, the authors adopt the idea of B-splines. Each $\phi(x)$ is represented as:

\begin{equation}\label{eq:4}
\phi(x)=w_b b(x)+w_s \operatorname{spline}(x)
\end{equation}
where $w_b$ and $w_s$ are both trainable scalar parameters, and $b(x)$ is defined as:
\begin{equation}
    b(x)=\operatorname{silu}(x)=x /\left(1+e^{-x}\right)
\end{equation}
$b(x)$ plays a role similar to residual connections.
The $\operatorname{spline}(x)$ component is parameterized as a linear combination of B-splines:
\begin{equation}\label{eq:6}
    \operatorname{spline}(x)=\sum_i c_i B_i(x)
\end{equation}
where $c_i$ are trainable parameters, and $B_i(x)$ are the B-spline basis functions.

\paragraph{Application} KAN has demonstrated significant success in various domains, particularly in tasks involving high-dimensional data and complex function approximation. For instance, KANs have been effectively applied in areas such as image and video processing~\cite{VisionKAN}, as well as function approximation~\cite{Neuromancer2023}. In these contexts, KANs have been utilized to model intricate processes by decomposing high-dimensional relationships into manageable, low-dimensional representations.

However, the application of KAN to MetaBBO methods remains an area of active exploration. One related study~\cite{kan-ea} employs KAN within a Surrogate-Assisted Evolutionary Algorithm (SAEA), where KAN serves as a surrogate model to guide the selection of promising solutions during the search process.

\begin{figure}[t]
    \centering
    \includegraphics[width=0.9\columnwidth]{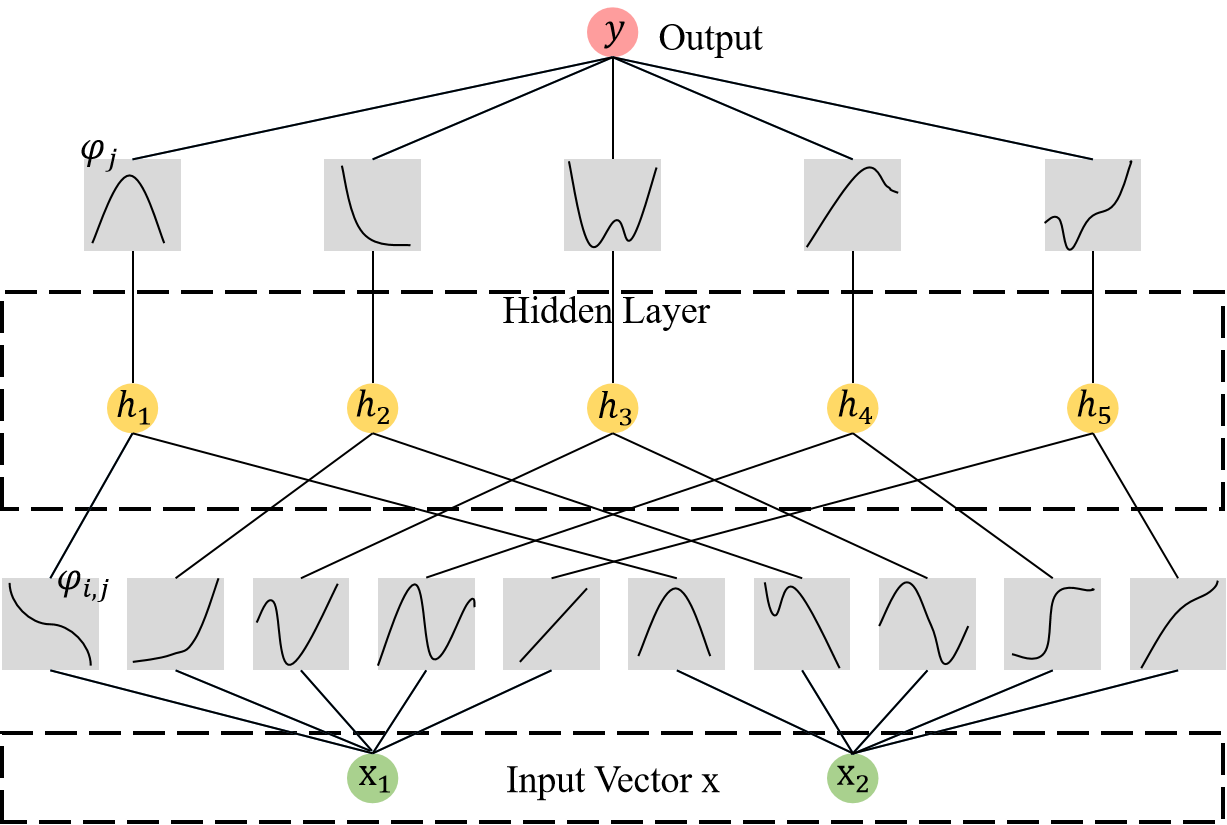}
    \caption{Neural network architecture of KAN used in this paper, where input and hidden layer dimension is limited to $2$ and $5$ respectively for the convenience of presentation.}
    \label{fig:kan} 
    \vspace{-3mm}
\end{figure}

\section{Methodology}
As we elaborated in previous sections, a key bottleneck of existing MetaBBO approaches is that they unavoidably consume massive number of function evaluations to train the meta-level policy, which might be neither practical nor efficient in many realistic BBO scenarios. The core motivation of Surr-RLDE is to explore an interesting research question: Can MetaBBO be meta-trained with surrogate models of the low-level optimization problems~(which might introduces biased learning signal), while maintaining the learning effectiveness and the final optimization performance? To find the answer, we design two learning stages: Surrogate Learning Stage~(SLS) and Policy Learning Stage~(PLS) in Surr-RLDE, of which the problem definitions and detailed technical designs are elaborated in Section~\ref{sec3.1} and Section~\ref{sec3.2} respectively. 
\subsection{Surrogate Learning Stage}\label{sec3.1}
\subsubsection{Problem Definition}
The first stage of Surr-RLDE is Surrogate Learning Stage~(SLS). SLS aims to approximate the objective functions of the low-level BBO problems used to meta-train the meta-level policy, hence saving function evaluations on these problems during the training. Intuitively, the primary objective in this stage is to learn a surrogate model $F_\theta$~(a neural network parameterized with $\theta$) that regresses a given $D$-dimensional BBO problem $f: \mathbb{R}^D \rightarrow \mathbb{R}$ with minimal MSE regression error~\cite{surrogate_regression_ddea}:
\begin{equation}\label{eq:7}
    \mathrm{Minimize:} \frac{1}{N} \sum_{i=1}^{N} \frac{1}{2}(f(x_i) -F_\theta(x_i))^2, x_i \in D_f
\end{equation}
where $D_f: \{x_i, f(x_i)\}_{i=1}^N$ is a collection of $N$ solutions uniformly and randomly sampled from the definition domain of the decision variables in BBO problem $f$, and their corresponding function values. In surrogate learning literature~\cite{surrogate_survey1,surrogate_survey2}, $D_f$ is regarded as the training dataset and the above MSE regression error term is regarded as the training loss function to learn optimal $\theta^*$ of the surrogate model $F_\theta$. Once trained, $F_{\theta^*}$ is expected to serve as an accurate surrogate function on those solution positions out of $D_f$.  

\begin{algorithm}[t]
\caption{Surrogate Model Training Process.}
\label{alg:sls}\small
\KwIn{Dataset $D_f$, Initialized model $F_\theta$, Batch size $N_{batch}$, Training epochs $T_{mse}$ and $T_{roa}$, Learning rate $\eta$.}
\KwOut{Optimal surrogate model $F_{\theta^*}$.}
\For{$epoch \leftarrow 1$ \KwTo $T_{mse}$}{
    \For{each batch in $D_f$}{
        $\mathcal{L}(\theta) \leftarrow \frac{1}{N_{batch}}\sum_{i=1}^{N_{batch}} MSE(f(x_i), F_\theta(x_i))$; \\
        $\theta \leftarrow \theta - \eta \nabla_\theta \mathcal{L}(\theta)$; 
    }
}
\state{Set $\lambda$ in Eq.~\ref{eq:loss} as $1$;\\}
\For{$epoch \leftarrow 1$ \KwTo $T_{roa}$}{
    \For{each batch in $D_f$}{
        Sort $x_i$ in the batch by $f(x_i)$ in descending order;\\
        Compute ROA loss $\mathcal{L}(\theta)$ described in Eq.~\ref{eq:loss};\\
        $\theta \leftarrow \theta - \eta \nabla_\theta \mathcal{L}(\theta)$; 
    }
    $\lambda \leftarrow \lambda \cdot (1 - \frac{epoch}{T_{mix}})$
}
\KwRet{The trained $F_\theta$}
\end{algorithm}

\subsubsection{Surrogate Model Design}
In Surr-RLDE, we employ a KAN~\cite{liu2024kan} neural network for SLS. We illustrate a simple architecture example of KAN~(with a hidden layer) in Fig.~\ref{fig:kan}, where the input dimension $L_{in}$ is $2$, the hidden dimension $L_{hidden}$ is $5$, the output dimension $L_{out}$ is $1$. The computational workflow is that: first, given an input $x$ with $L_{in}$ dimensions, for each $i$-th dimension $x_i$, we connect it with each $j$-th dimension $h_j$ in the hidden layer, through a transformation function $\phi_{i,j}$~(see Eq.~\ref{eq:4}). The final post-activation value of each hidden unit $h_j$ is the summation of all connections with the input $x$: $h_j = \sum_{i=1}^{L_{in}}\phi_{i,j}(x_i)$. To attain the output value $y$, we further connect each $h_j$ in the hidden layer with it, through a transformation function $\phi_{j}$. Then $y$ is attained by summing up all connections: $y = \sum_{j=1}^{L_{hidden}}\phi_{j}(h_j)$. In the rest of this paper, we use $F_\theta$ to refer to the KAN-based surrogate model, where $\theta$ is the trainable parameters in the model, including the combination weights~($w_b$, $w_s$ and $c_k$ in Eq.~\ref{eq:4} and Eq.~\ref{eq:6}) within the transformation function $\phi$ of each connections. Given an input $x$, we attain $y = F_\theta(x)$ directly. 

We have to note that there are two core insights that motivate us to use KAN as the surrogate model rather than MLP neural network: a) KAN replaces the fixed activation functions in MLP by learnable spline function combinations, while maintaining the fully-connected structures. This helps KAN learn both the generalized compositional structure of an optimization problem and the non-linearity within each univariate function components. b) Some recent studies~\cite{kan-application1,kan-application2} indicate that KAN often outperforms MLP in diverse machine learning tasks, considering both accuracy and efficiency.   

\begin{figure}[t]
    \centering
    \includegraphics[width=0.7\columnwidth]{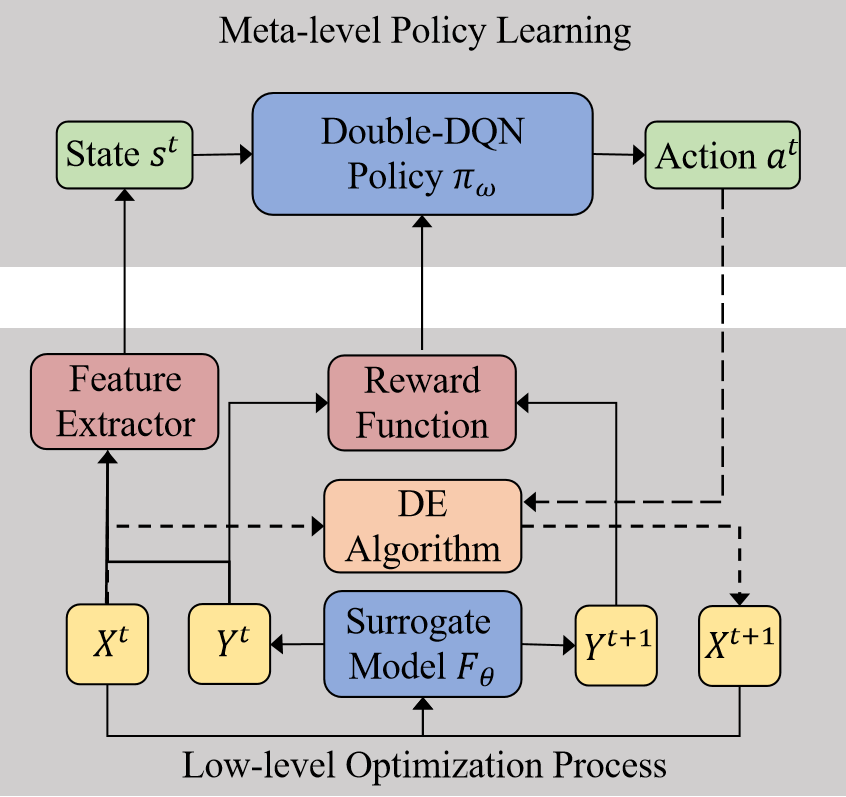}
    \caption{Workflow of Policy Learning Stage in Surr-RLDE.}
    \label{fig:workflow} 
    \vspace{-4mm}
\end{figure}

\subsubsection{Relative-Order-Aware Loss}
During our preliminary experimental study, we found that naively training the proposed KAN network by the MSE regression loss function~(see Eq.~\ref{eq:7}) on the training dataset $D_f$ can not ensure stability of the subsequent MetaBBO policy learning stage. We delve into this observation to explore the tricky part behind, and found that the learning failure comes from the regression process itself. For a complex function $f$ with diverse and intricate local landscapes~\cite{ela_original}, the regression loss is hard to be minimized to $0$, which leads to the relative orders of the predicted function values of some data points in $D_f$ violate the ground truth of their function values on $f$, which is also observed in recent surrogate learning studies~\cite{l2rank, clddea}. We provide an example here: suppose there are five data points in $D_f$, which are $[x_1,x_2,x_3,x_4,x_5]$, sorted by descending order of their true objective values $f(x)$: $[9,5.1,5,3,1]$. However, after training our KAN-based surrogate model $F_\theta$, the predicted objective values $F_\theta(x)$ are $[9, 5, 5.1, 3, 1]$. In this case, although $F_\theta$ achieves ideal accuracy on most of the data points, it misjudges the relative order of $x_2$ and $x_3$. Although such misjudgment might not interfere classical machine learning tasks, it is indeed problematic for MetaBBO approaches. As we introduced in Section~\ref{sec:2.1}, in MetaBBO, the training of its meta-level policy relies on the correct feedback from the low-level optimization process which indicates the relative performance improvement. In the case that we substitute the original optimization problem $f$ by our trained KAN model $F_\theta$, the low-level optimization process might return an incorrect relative performance improvement due to the misjudgment of the relative order of two solution points. Apparently, this might lead to potential training failure.   

To address the risk of misjudgment of the relative order, we propose a novel Relative-Order-Aware~(ROA) loss function, which combines the strength of the MSE regression loss to improve the model's fitting accuracy and an order correction~(OC) term that forces $F_\theta$ to maintain the ground truth of the relative orders among training data. Given a batch of $N_{batch}$ input data points sampled from $D_f$ and sorted by their actual objective values in descending order, the proposed ROA loss function is formulated as below:
\begin{equation}\label{eq:loss}
\begin{split}
    \mathcal{L}(\theta) = \frac{1}{N_{batch}}\sum_{i=1}^{N_{batch}} \left[ \lambda \cdot MSE(f(x_i), F_\theta(x_i)) + OC(F_\theta(x_i))\right]\\
    OC(F_\theta(x_i)) = \frac{1}{2}\left[|f(x_{i-1}) - F_\theta(x_i)| + |F_\theta(x_i) - f(x_{i+1})|\right]
\end{split}
\end{equation}
where the total loss is a weighted combination of the original MSE loss and the OC term. For the $i$-th data point $x_i$ in the sorted data batch, we force the prediction value of the surrogate model, say $F_\theta(x_i)$, to locate between the ground truth objective values of $x_{i-1}$ and $x_{i+1}$. The OC term is minimized only when the relative orders of the predicted values in the data batch is consistent with the ground truth value. 

We present the overall training workflow of SLS in Algorithm~\ref{alg:sls}. It first goes through a pre-training phase where the loss function is the naive MSE~(line $1\sim 6$). This phase aims to let the model be aware of the overall structure of the target function $f$. Then it goes through the second phase, where the loss function is our proposed ROA loss in Eq.~\ref{eq:loss}~(line $7 \sim 14$). The combination weights $\lambda$ in this phase is initially set to $1$ and then decay linearly, which strikes a good balance in learning both the prediction accuracy and the relative order accuracy.

\subsection{Policy Learning Stage}\label{sec3.2}
Our proposed Policy Learning Stage~(PLS) mirrors existing MetaBBO methods~\cite{deddqn,dedqn,gleet,guo2024configx} which leverage RL~\cite{sutton2018reinforcement} techniques for dynamic algorithm configuration~(DAC) in Evolutionary Algorithms~(EAs). Since this paper serves as a preliminary study on the positive effects of surrogate learning techniques in MetaBBO, we define a relatively simple MetaBBO task as the experimental operation basis, which is illustrated in Figure~\ref{fig:workflow}. 

Specifically, a Double-DQN~\cite{double-dqn} agent is employed in the meta-level to serve as the meta-level policy $\pi_\omega$, dictating flexible mutation operator configurations for the low-level DE algorithm dynamically. In the low-level optimization, given an optimization problem distribution $\mathbb{P}$, we first sample a collection of training problem instances $\mathbb{P}_{train}$ used for policy learning. For each instance $f\in \mathbb{P}_{train}$, we follow the procedure in SLS and train a surrogate model $F_\theta$. Then these surrogate models are used as the substitution of $\mathbb{P}_{train}$ in the low-level optimization. We have to note that the overall workflow of PLS in Surr-RLDE shows consistency with what we have elaborated in Section~\ref{sec:2.1}. The only difference is that we replace the evaluation function in the low-level optimization by surrogate models, which plays two key roles in Surr-RLDE: 

a) \textbf{Feature Extraction}: at each time step $t$ of the low-level optimization, the population of the DE algorithm, $X^t$, is fed into surrogate model $F_\theta$ to obtain the objective values $Y^t$. Then a feature extractor module processes $X^t$ and $Y^t$ into the optimization state $s^t$ in this time step. To ensure an accurate profiling of the current optimization state, the surrogate model need to be as accurate as possible, compared with the original function $f$. 

b) \textbf{Reward Computation}: Once the meta-level policy $\pi_\omega$ receives $s^t$, it accordingly outputs an action $a^t$~(in Surr-RLDE, $a^t$ denotes different mutation operator configurations) for the low-level DE. Then DE algorithm load the dictated mutation configuration $a^t$ and evolves $X^t$ towards next generation $X^{t+1}$, with corresponding $Y^{t+1}$ evaluated by $F_\theta$. A reward signal in computed by a reward function, according to the relative performance improvement between $Y^{t+1}$ and $Y^{t}$. Then the meta-level policy $\pi_\omega$ is trained to maximize the accumulated rewards along the low-level optimization process. To ensure the reward signal accurately reflects the effects of $\pi_\omega$, the objective values predicted by $F_\theta$ should at least keep the relative order consistency. This is exactly the motivation of our proposed ROA loss. 

We have to highlight here that integrating surrogate learning into MetaBBO's learning system marks a significant step forward for this domain. Intuitively, existing MetaBBO approaches often train their meta-policy for hundreds of epochs and over dozens of training problem instances, which implies massive function evaluations. In surr-RLDE, it only consumes a small number of function evaluations to construct the training dataset $D_f$ for training the surrogate model, which greatly reduces evaluation cost. In subsequent sections, we elaborate some specific designs in PLS of Surr-RLDE.  

\begin{algorithm}[t]
\caption{Policy Learning in Surr-RLDE.}
\label{alg:pls}\small
\KwIn{Surrogate models $\{F_\theta^{(k)}\}_{k=1}^{|P_{train}|}$, Initialized prediction policy $\pi_\omega$, Maximum learning steps $maxLS$, Maximum function evaluations $maxFEs$ per optimization run, Update period $G_{up}$ of target policy, $DE$ algorithm.}
\KwOut{Optimal meta-level policy $\pi_{\omega^*}$.}
\state{Initialize replay buffer $RM \leftarrow \emptyset$;\\}
\state{Copy parameter weights in $\pi_\omega$ to taget policy $\pi_{target}$;\\}
\state{Initialize learning step $LS \leftarrow 0$;\\}
\While{$LS \leq maxLS$}{
    \For{$k \leftarrow 1$ \KwTo $|P_{train}|$}{
        \state{initialize generation flag $t \leftarrow 0$;\\}
        \state{$X^t \leftarrow DE.initialize()$, $Y^t = F_\theta^{(k)}(X^t)$;\\}
        \state{$FEs \leftarrow DE.population\_size$;\\}
        \state{Extract optimization state $s^t$ following Section~\ref{sec:3.2.1};\\}
        \While{$FEs \leq maxFEs$}{
            \state{$a^t \leftarrow \pi_\omega(s^t).epsilon\_greedy()$;\\}
            \state{Set $DE$'s mutation configuration as $a^t$;\\}
            \state{$X^{t+1} \leftarrow DE.step(X^t,Y^t)$, $Y^{t+1} \leftarrow F_\theta^{(k)}(X^{t+1})$;\\}
            \state{Compute reward $r^t$ following Section~\ref{sec:3.2.3};\\}
            \state{Extract optimization state $s^{t+1}$ following Section~\ref{sec:3.2.1};\\}
            \state{$RM.insert(\{s^t, a^t, r^t, s^{t+1}\})$;\\}
            \state{Update $\pi_\omega$ following Eq.~\ref{eq:qloss}, $LS \leftarrow LS +1$;\\}
            \If{$LS \% G_{up} = 0$}{
                \state{Update $\pi_{target}$ by $\pi_\omega$'s parameter weights;\\}
            }
            \state{$FEs \leftarrow FEs + DE.population\_size$, $t \leftarrow t+1$;\\}
        }
    }
}

\KwRet{The trained prediction policy $\pi_\omega$}

\end{algorithm}

\subsubsection{State Design}\label{sec:3.2.1}
In MetaBBO approaches that leverage RL as the techniques, the state representation $s^t$ should informatively reflect the dynamic optimization status of the low-level optimization process, so as to ensure the learning effectiveness of the meta-level policy. In BBO, a common practice for featuring optimization information is Exploratory Landscape Analysis~(ELA)~\cite{ela_original}, which includes six feature groups of $55$ single features, each profiling a particular aspect of the optimization problem such as local-optimum properties, fitness correlation distances, convexity, etc. However, computing some of these features for each low-level optimization step in MetaBBO is impractical due to the computational complexity and requirement for large scale sampling. To address this issue, existing MetaBBO approaches lean to select an easy-to-compute subset from ELA features~\cite{dedqn}. A recent work SYMBOL~\cite{chen2024symbol}, proposed a novel $9$-dimensional state design that shows both effectiveness and efficiency for MetaBBO. It comprises three parts: a) $s_1 \sim s_3$ compute the distributional properties of the solution population. b) $s_4\sim s_6$ are a subset of fitness distance correlation features in ELA, reflecting the optimization convergency. c) $s_7 \sim s_9$ are some time-stamp features that indicate the pre-mature and evolution progress. We borrow this idea from SYMBOL to serve as the feature extraction module in Surr-RLDE. For detailed definition of these features, refer to the original paper. 
\subsubsection{Action Space Design}\label{sec:3.2.2}
The action space in PLS of Surr-RLDE is a series of mutation operator configurations which combines various mutation operators and various mutation strength values. For mutation operator, we select five well-known variants: DE/rand1, DE/best1, DE/current-to-rand, DE/current-to-pbest and DE/current-to-best, which show diverse exploration-exploitation behavior. For mutation strength $F$, we provide three options: $0.1$, $0.5$ and $0.9$. The action space $A$ hence comprises $5 \times 3 = 15$ optional actions, each featuring a mutation variant and corresponding mutation strength. By dynamically configuring the low-level DE algorithm with desired mutation configuration at each optimization step, Surr-RLDE could improve the overall optimization performance of DE with single mutation configuration. We note that the choice of the DE algorithm is based on its common use in existing MetaBBO approaches and empirical considerations.

\subsubsection{Reward Design}\label{sec:3.2.3}
We adopt a simple and classic reward function in PLS of Surr-RLDE:
\begin{equation}\label{eq:reward}
    r^t = \begin{cases}
        0, \quad if \quad y^{*,t+1} \leq y^{*,t}\\
        1, \quad otherwise
    \end{cases}
\end{equation}
where $y^{*,t}$ denotes the objective value of the best solution found so far, evaluated by the surrogate model $F_\theta$. A positive reward signal is returned to the meta-level policy if a better solution is found. This reward function encourages the meta-level policy to dictate desired mutation configuration for the low-level DE algorithm, hence promoting the overall optimization progress.

\subsubsection{Learning Paradigm}
The overall learning paradigm of PLS in Surr-RLDE is presented in Algorithm~\ref{alg:pls}. We maintain a Double-DQN~\cite{double-dqn} agent as the meta-policy, which includes two neural networks: prediction policy $\pi_\omega$ and target policy $\pi_{target}$. The prediction policy is used to sample mutation configurations~(line $11$) for the low-level DE algorithm, following an $epsilon\_greedy$ strategy, where either the action with maximum predicted Q-values is sampled or the action in randomly chosen, with a small probability. The target policy $\pi_{target}$ serve as an invariant baseline, allowing for the removal of positive bias in regressing the Q-values of the prediction policy $\pi_\omega$. During the low-level optimization~(line $10\sim 22$), we use surrogate models of the training problem set $P_{train}$ as the objective functions to evaluate solution population~(line $13$). For each training step~(line $17$), we first sample a batch of transitions from a replay memory $RM$. Then for each transition $\{s^t, a^t, r^t, s^{t+1}\}$ the Bellman loss function used for updating the policy $\pi_\omega$ is formulated as below:
\begin{equation}\label{eq:qloss}
    \mathcal{L}(\omega) = \frac{1}{2}\left[[r^t +\gamma \mathop{\arg\max}\limits_{a \in A}\pi_{target}(s^{t+1},a)] -\pi_\omega(s^t,a^t)\right]^2
\end{equation}
where we use Q-values of the target policy as the label to regress Q-values of $\pi_\omega$. The training process lasts for $maxLS$ learning steps and covers all training problem instances in $P_{train}$. The target policy is updated by the parameter weights of the up-to-date prediction policy every $G_{up}$ learning steps~(line $18\sim 20$). Once training ends, the learned prediction policy $\pi_\omega$ could be directly applied for configuring DE algorithm on solving unseen optimization problems. 

\begin{table*}[t]
\caption{Generalization performance of the trained Surr-RLDE and baselines on test set $P_{test}$}
\label{tab:IID}
\resizebox{0.9\textwidth}{!}{%
\begin{tabular}{c|cc|cc|cc|cc|cc|cc|cc|cc|c}
\hline
                               & \multicolumn{2}{c|}{Weierstrass} &\multicolumn{2}{c|}{Schaffers}       & \multicolumn{2}{c|}{Schaffers\_high\_cond}        & \multicolumn{2}{c|}{Composite\_Grie\_rosen}        & \multicolumn{2}{c|}{Gallagher\_101Peaks}        & \multicolumn{2}{c|}{Gallagher\_21Peaks}        & \multicolumn{2}{c|}{Katsuura}        & \multicolumn{2}{c|}{Lunacek\_bi\_Rastrigin}        & \multirow{3}{*}{Avg Rank} \\ \cline{1-17}
                               & mean      & \multirow{2}{*}{rank} & mean      & \multicolumn{1}{c|}{\multirow{2}{*}{rank}} & mean      & \multirow{2}{*}{rank} & mean      & \multirow{2}{*}{rank} & mean      & \multirow{2}{*}{rank} & mean      & \multirow{2}{*}{rank} & mean      & \multirow{2}{*}{rank} & mean      & \multirow{2}{*}{rank} &                           \\
                               & (std)       &                       & (std)       & \multicolumn{1}{c|}{}                      & (std)       &                       & (std)       &                       & (std)       &                       & (std)       &                       & (std)       &                       & (std)       &                       &                           \\ \hline
\multirow{2}{*}{Surr-RLDE} & 6.914E+00 & \multirow{2}{*}{5}    & 4.330E-02 & \multirow{2}{*}{3}                         & 2.232E-01 & \multirow{2}{*}{3}    & 1.769E+00 & \multirow{2}{*}{3}    & 3.448E-07 & \multirow{2}{*}{1}    & 1.972E-05 & \multirow{2}{*}{1}    & 1.458E+00 & \multirow{2}{*}{4}    & 4.245E+01 & \multirow{2}{*}{4}    & \multirow{2}{*}{3}        \\
                               & \small($\pm$1.603E+00) &                       & \small($\pm$3.622E-02) &                                            & \small($\pm$1.007E-01) &                       & \small($\pm$3.189E-01) &                       & \small($\pm$1.071E-06) &                       & \small($\pm$1.277E-04) &                       & \small($\pm$3.026E-01) &                       & \small($\pm$9.243E+00) &                       &                           \\\hline
\multirow{2}{*}{Surr-RLDE-O}   & 2.173E+00 & \multirow{2}{*}{2}    & 8.484E-02 & \multirow{2}{*}{4}                         & 4.270E-01 & \multirow{2}{*}{4}    & 1.427E+00 & \multirow{2}{*}{2}    & 1.719E+00 & \multirow{2}{*}{5}    & 2.908E+00 & \multirow{2}{*}{3}    & 1.417E+00 & \multirow{2}{*}{3}    & 4.525E+01 & \multirow{2}{*}{5}    & \multirow{2}{*}{3.5}    \\
                               & \small($\pm$2.443E+00) &                       & \small($\pm$5.639E-02) &                                            & \small($\pm$2.730E-01) &                       & \small($\pm$4.405E-01) &                       & \small($\pm$1.812E+00) &                       & \small($\pm$3.250E+00) &                       & \small($\pm$2.865E-01) &                       & \small($\pm$7.022E+00) &                       &                           \\\hline
\multirow{2}{*}{DE\_DDQN}      & 4.173E+00 & \multirow{2}{*}{3}    & 2.424E-04 & \multirow{2}{*}{1}                         & 8.645E-04 & \multirow{2}{*}{1}    & 2.336E+00 & \multirow{2}{*}{5}    & 7.973E-01 & \multirow{2}{*}{4}    & 3.530E+00 & \multirow{2}{*}{4}    & 1.479E+00 & \multirow{2}{*}{6}    & 4.014E+01 & \multirow{2}{*}{3}    & \multirow{2}{*}{3.375}    \\
                               & \small($\pm$2.292E+00) &                       & \small($\pm$3.196E-04) &                                            & \small($\pm$8.104E-04) &                       & \small($\pm$4.484E-01) &                       & \small($\pm$1.769E+00) &                       & \small($\pm$2.863E+00) &                       & \small($\pm$2.262E-01) &                       & \small($\pm$6.514E+00) &                       &                           \\\hline
\multirow{2}{*}{DE\_DQN}       & 2.108E+01 & \multirow{2}{*}{9}    & 7.181E+00 & \multirow{2}{*}{9}                         & 2.550E+01 & \multirow{2}{*}{9}    & 1.226E+01 & \multirow{2}{*}{9}    & 4.484E+01 & \multirow{2}{*}{9}    & 5.847E+01 & \multirow{2}{*}{9}    & 3.473E+00 & \multirow{2}{*}{9}    & 1.586E+02 & \multirow{2}{*}{9}    & \multirow{2}{*}{9}        \\
                               & \small($\pm$4.650E+00) &                       & \small($\pm$1.265E+00) &                                            & \small($\pm$4.107E+00) &                       & \small($\pm$2.320E+00) &                       & \small($\pm$1.103E+01) &                       & \small($\pm$9.404E+00) &                       & \small($\pm$8.317E-01) &                       & \small($\pm$2.128E+01) &                       &                           \\\hline
\multirow{2}{*}{GLEET}         & 4.944E-01 & \multirow{2}{*}{1}    & 1.031E-01 & \multirow{2}{*}{5}                         & 4.965E-01 & \multirow{2}{*}{5}    & 1.801E+00 & \multirow{2}{*}{4}    & 2.330E-01 & \multirow{2}{*}{3}    & 9.022E-01 & \multirow{2}{*}{2}    & 1.216E+00 & \multirow{2}{*}{1}    & 2.995E+01 & \multirow{2}{*}{1}    & \multirow{2}{*}{2.75}    \\
                               & \small($\pm$4.408E-01) &                       & \small($\pm$7.516E-02) &                                            & \small($\pm$4.083E-01) &                       & \small($\pm$5.846E-01) &                       & \small($\pm$6.705E-01) &                       & \small($\pm$2.035E+00) &                       & \small($\pm$3.707E-01) &                       & \small($\pm$9.255E+00) &                       &                           \\\hline
\multirow{2}{*}{DE}            & 6.996E+00 & \multirow{2}{*}{6}    & 2.438E-03 & \multirow{2}{*}{2}                         & 3.110E-02 & \multirow{2}{*}{2}    & 1.109E+00 & \multirow{2}{*}{1}    & 8.647E-02 & \multirow{2}{*}{2}    & 5.600E+00 & \multirow{2}{*}{6}    & 1.554E+00 & \multirow{2}{*}{8}    & 3.915E+01 & \multirow{2}{*}{2}    & \multirow{2}{*}{3.625}    \\
                               & \small($\pm$1.921E+00) &                       & \small($\pm$9.000E-03) &                                            & \small($\pm$4.503E-02) &                       & \small($\pm$3.204E-01) &                       & \small($\pm$3.902E-01) &                       & \small($\pm$5.354E+00) &                       & \small($\pm$2.255E-01) &                       & \small($\pm$4.472E+00) &                       &                           \\\hline
\multirow{2}{*}{PSO}           & 6.439E+00 & \multirow{2}{*}{4}    & 1.533E+00 & \multirow{2}{*}{6}                         & 5.630E+00 & \multirow{2}{*}{6}    & 3.104E+00 & \multirow{2}{*}{6}    & 1.954E+00 & \multirow{2}{*}{6}    & 4.513E+00 & \multirow{2}{*}{5}    & 1.460E+00 & \multirow{2}{*}{5}    & 5.948E+01 & \multirow{2}{*}{6}    & \multirow{2}{*}{5.5}      \\
                               & \small($\pm$1.817E+00) &                       & \small($\pm$2.072E-01) &                                            & \small($\pm$9.901E-01) &                       & \small($\pm$6.871E-01) &                       & \small($\pm$2.664E+00) &                       & \small($\pm$4.161E+00) &                       & \small($\pm$2.711E-01) &                       & \small($\pm$7.078E+00) &                       &                           \\\hline
\multirow{2}{*}{SAHLPSO}       & 8.047E+00 & \multirow{2}{*}{7}    & 2.281E+00 & \multirow{2}{*}{7}                         & 8.243E+00 & \multirow{2}{*}{7}    & 5.267E+00 & \multirow{2}{*}{7}    & 2.274E+00 & \multirow{2}{*}{7}    & 6.643E+00 & \multirow{2}{*}{7}    & 1.380E+00 & \multirow{2}{*}{2}    & 8.842E+01 & \multirow{2}{*}{7}    & \multirow{2}{*}{6.375}    \\
                               & \small($\pm$5.463E+00) &                       & \small($\pm$1.006E+00) &                                            & \small($\pm$3.467E+00) &                       & \small($\pm$1.184E+00) &                       & \small($\pm$4.457E+00) &                       & \small($\pm$8.720E+00) &                       & \small($\pm$5.432E-01) &                       & \small($\pm$1.761E+01) &                       &                           \\\hline
\multirow{2}{*}{Random Search} & 1.005E+01 & \multirow{2}{*}{8}    & 3.974E+00 & \multirow{2}{*}{8}                         & 1.406E+01 & \multirow{2}{*}{8}    & 6.592E+00 & \multirow{2}{*}{8}    & 1.330E+01 & \multirow{2}{*}{8}    & 1.753E+01 & \multirow{2}{*}{8}    & 1.482E+00 & \multirow{2}{*}{7}    & 1.081E+02 & \multirow{2}{*}{8}    & \multirow{2}{*}{7.875}    \\
                               & \small($\pm$1.403E+00) &                       & \small($\pm$5.344E-01) &                                            & \small($\pm$2.263E+00) &                       & \small($\pm$9.522E-01) &                       & \small($\pm$3.896E+00) &                       & \small($\pm$5.886E+00) &                       & \small($\pm$2.483E-01) &                       & \small($\pm$9.746E+00) &                       &                           \\ \hline
\end{tabular}
}
\end{table*}

\begin{figure*}[t]
\centering

\subfigure[\emph{Gallagher 21Peaks} 10D-S/R]{
\includegraphics[width=0.19\linewidth]{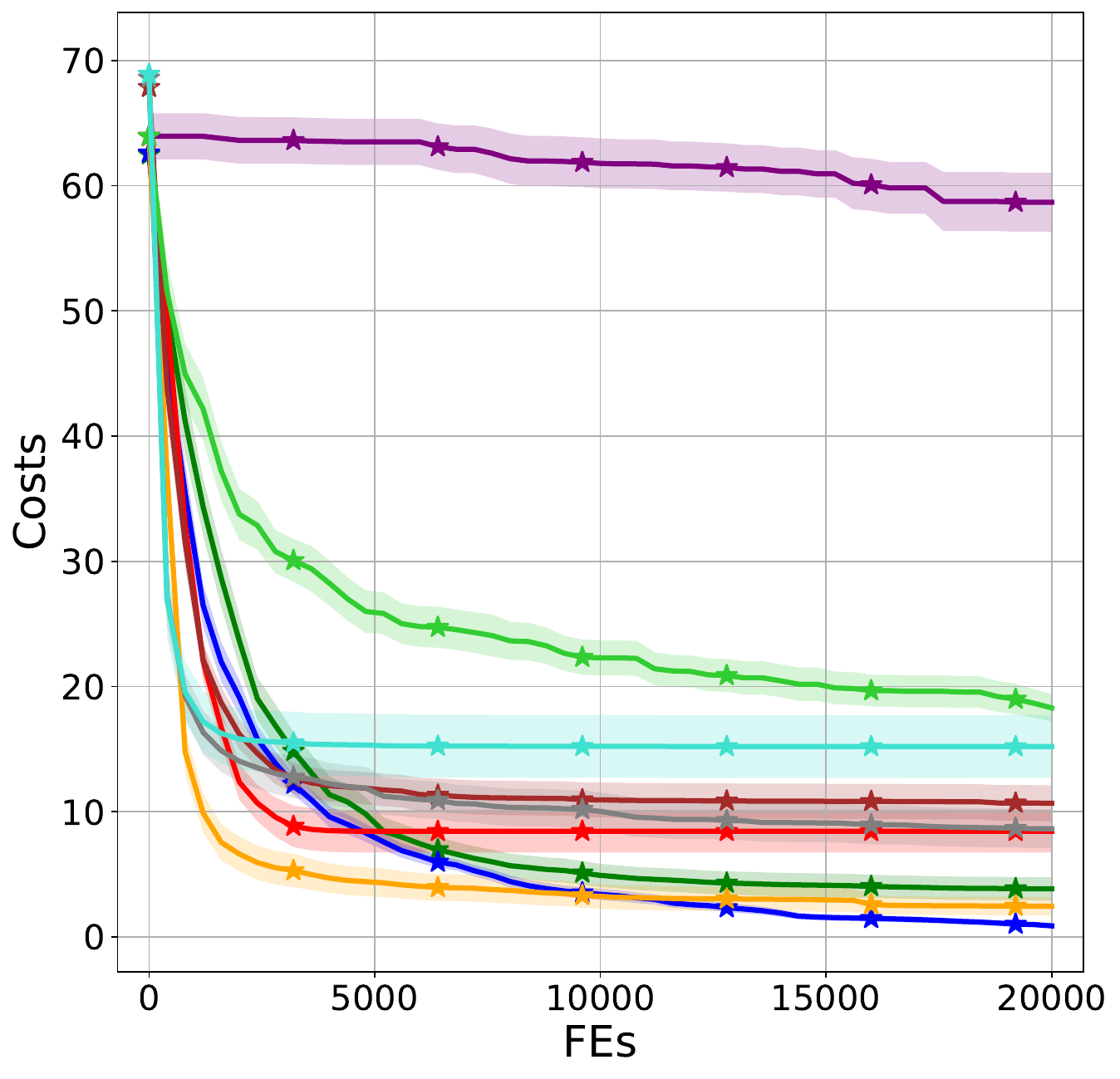}
}
\subfigure[\emph{Gallagher 101Peaks} 10D-S/R]{
\includegraphics[width=0.19\linewidth]{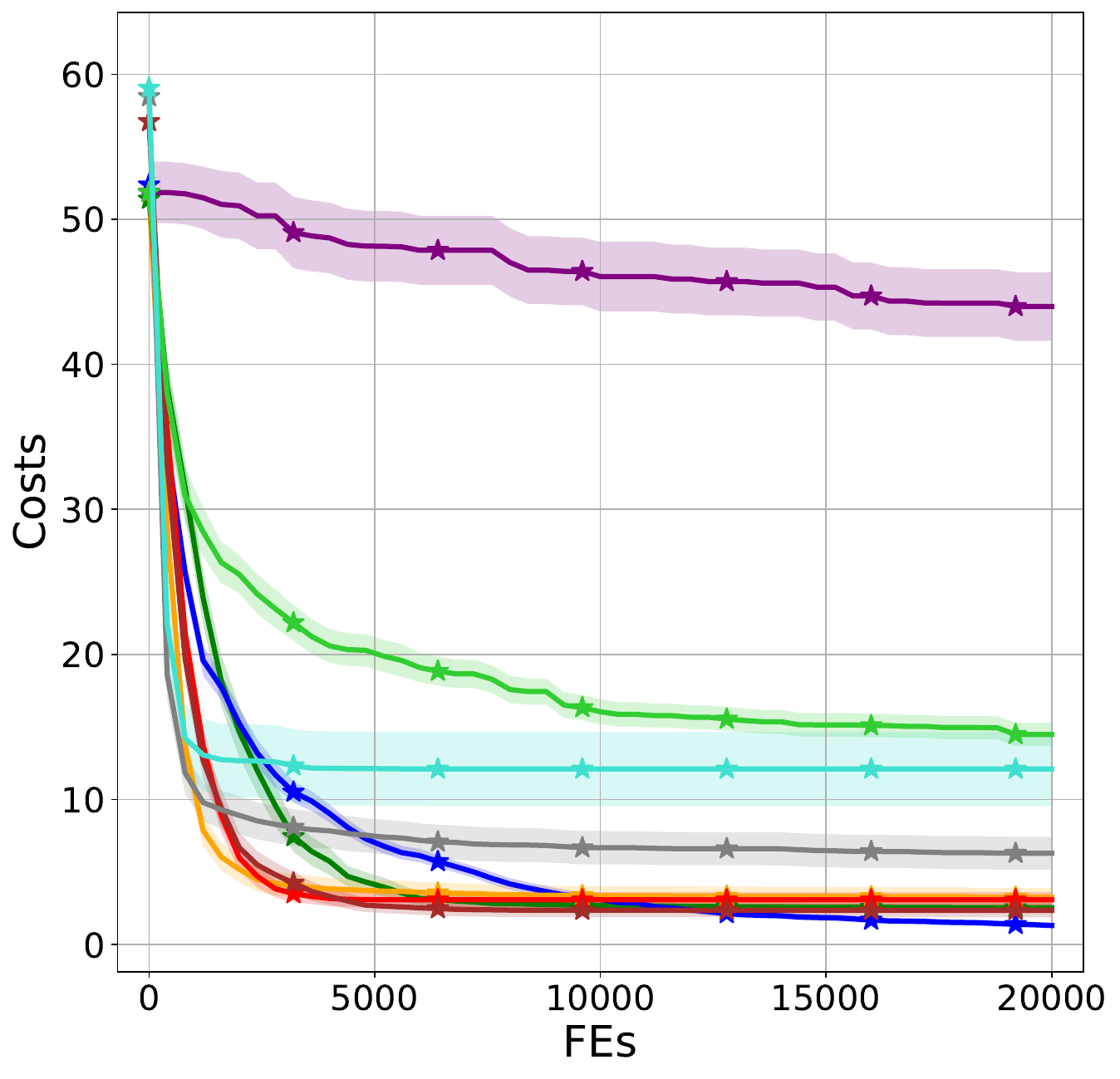}
}
\subfigure[\emph{Gallagher 21Peaks} 30D]{
\includegraphics[width=0.19\linewidth]{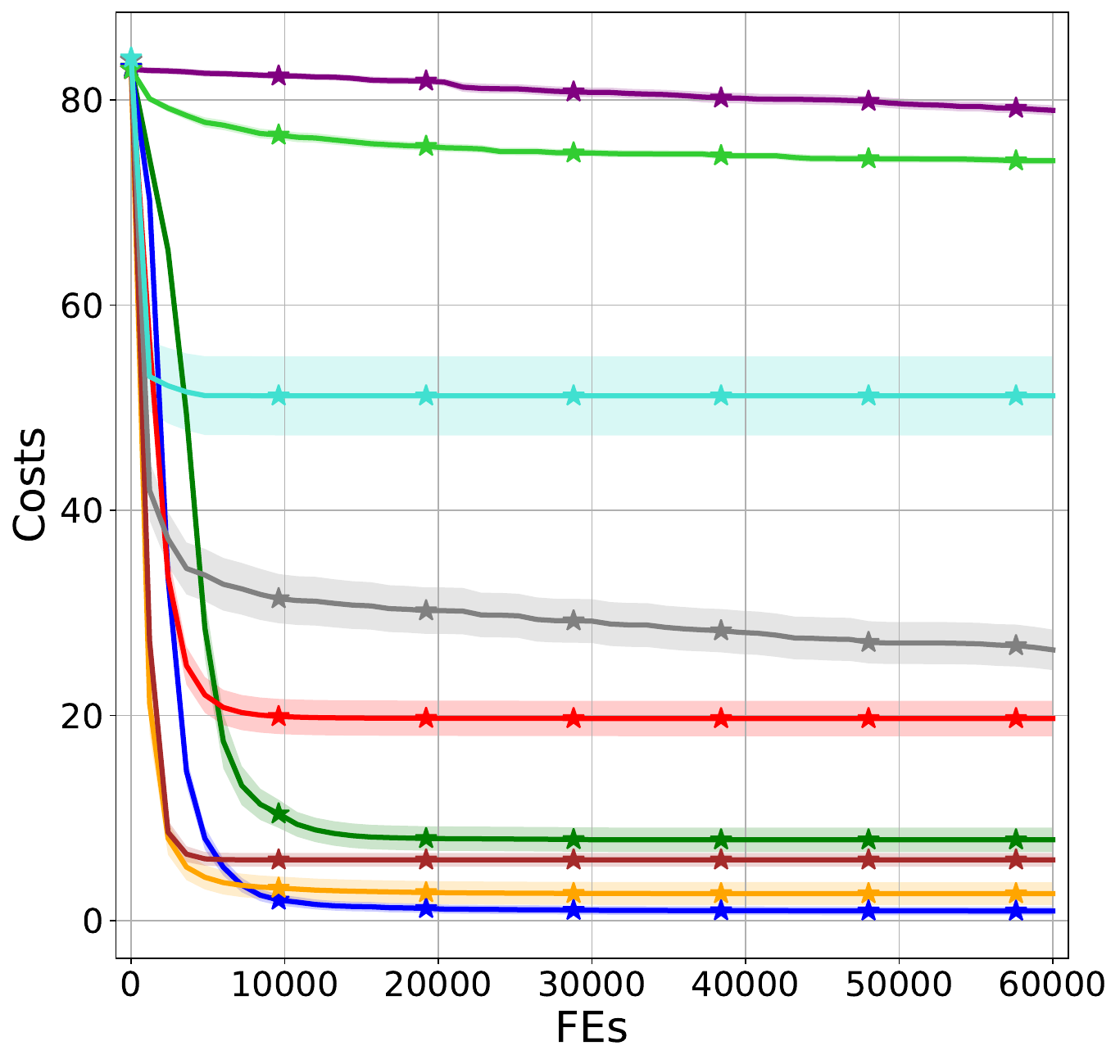}
}
\subfigure[\emph{Schaffers high cond} 30D]{
\includegraphics[width=0.19\linewidth]{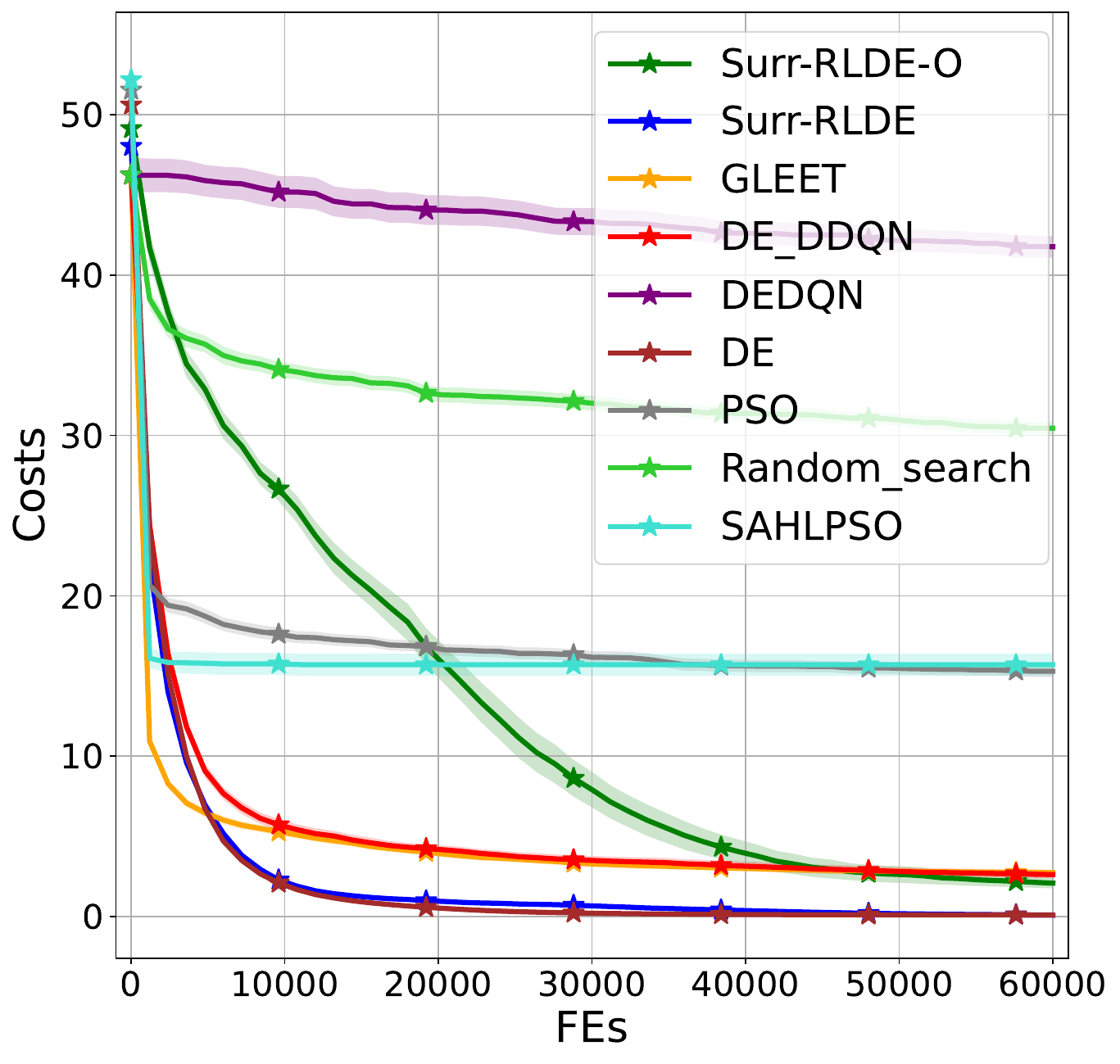}
}

\caption{Generalization performance of the trained Surr-RLDE and baselines on out-of-distribution optimization problems. }
\label{fig:OOD}
\vspace{-1em}
\end{figure*}
\section{Experimental Results}

\textbf{Optimization Problem Dataset.} In this paper, we adopt COCO-BBOB testsuites~\cite{bbob2010} as the basic optimization problem dataset, which contains 24 synthetic optimization problems with diverse properties including unimodal or multi-modal, separable or non-separable, adequate or weak global structure, etc. For train-test-split, we randomly select 16 problem instances for training and 8 for testing. Specifically, training set $P_{train}$ includes following functions: \textit{Sphere, Ellipsoidal, Rastrigin, BucheRastrigin, LinearSlope, AttractiveSector, StepEllipsoidal, Rosenbrockoriginal, Rosenbrockrotated, Ellipsoidalhighcond, Discus, BentCigar, SharpRidge, DifferentPowers, RastriginF15,Schwefel}; test set $P_{test}$ includes following functions: \textit{Weierstrass, Schaffers, Schaffershighcond, CompositeGrierosen, Gallagher101Peaks, Gallagher21Peaks, Katsuura, LunacekbiRastrigin}. Problem dimension is set to 10, searching range locates in $\left[-5, 5\right]$. For each instance,the $maxFEs$ is set to $2\times10^4$ and the random shift/rotation option is turned off.  

\textbf{Surrogate learning stage.} For this learning stage, we follow Algorithm~\ref{alg:sls} to train a separate KAN model for each problem instance in $P_{train}$. The KAN model is implemented following \texttt{pykan}\footnote{https://github.com/KindXiaoming/pykan}, with the layer configuration as $[10,10,1]$, grid size $G=5$ and B-spline order $k=5$. For each $f \in P_{train}$, we leverage Latin Hypercube Sampling (LHS) to sample $5 \times 10^4$ points and accordingly construct training data $D_f$ with these sample points and corresponding objective values. During training, batch size $N_{batch} = 100$, training epochs $T_{mse} = 300, T_{roa} = 1000$, learning rate $\eta = 0.01$. After training, we obtain $16$ models, each corresponds to the surrogate model of a problem instance in $P_{train}$.

\textbf{Policy learning stage.} We follow the algorithm template in MetaBox\footnote{https://github.com/GMC-DRL/MetaBox} to develop policy learning pipeline in Algorithm~\ref{alg:pls}. We built up a MLP network as the prediction policy $\pi_\omega$ in Double-DQN, specifically with 9D inputs, 15D outputs, 3 hidden layers with $\left[32, 64, 32\right]$ hidden neurons and relu activations. During training, maximum learning steps $maxLS$ is set to $1.5 \times 10^6$, update period $G_{up}$ for $\pi_{target}$ is set to 1000, discount factor $\gamma=0.99$, learning rate is set to $1e-4$. We utilize vanilla DE as the low-level optimizer. Within our implementation of DE, initialization strategy is set to uniformly sampling, crossover strategy is set to binary crossover with $Cr=0.7$.  

\textbf{Running platform.} All results presented are from a platform equipped with a RTX 2080Ti 11GB GPU, an Intel Xeon E5-2680 v4 @ 56x 3.3GHZ CPU and 128 GB RAM. 

\textbf{Baselines.} We compare Surr-RLDE with $7$ baselines, which can be divided into two categories: a) \textbf{BBO methods}: Random Search, DE~\cite{de_original}, PSO~\cite{pso_original}, SAHLPSO~\cite{sahlpso}. b) \textbf{MetaBBO methods}: DE-DDQN~\cite{deddqn}, DEDQN~\cite{dedqn} and GLEET~\cite{gleet}. DE-DDQN and DEDQN are two baselines using value-based RL techniques for dynamic mutation operator selection in DE. GLEET is an up-to-date baseline with strong performance using policy-gradient RL techniques for dynamic hyper-parameter tuning in PSO. In addition, we include two variants of Surr-RLDE in following experiments: a) \textbf{Surr-RLDE-O}, which skips the surrogate learning stage and uses the original training problems $P_{train}$ for function evaluation in policy learning stage, similar with existing MetaBBO's paradigm. b) \textbf{Surr-RLDE-MSE}, which sorely uses MSE loss~(Eq.~\ref{eq:7}) instead of our proposed ROA loss~(Eq.~\ref{eq:loss}) for the surrogate model training in SLS. All MetaBBO baselines follow their original settings in their papers, except that the training set and maximum learning steps coincide with the setting for Surr-RLDE, Surr-RLDE-O and Surr-RLDE-MSE, which is $P_{train}$ and $1.5\times 10^6$ respectively.


\subsection{Performance Comparison Results}
\subsubsection{In-Distribution Generalization}
After training our Surr-RLDE and Surr-RLDE-O, and other MetaBBO baselines DE-DDQN, DEDQN and GLEET on $P_{train}$, we test the learned meta-level policies in these baselines and the four BBO baselines on $P_{test}$, where the optimization problems hold the same dimension~(10) as the problems in $P_{train}$. We hence refer to this testing as the in-distribution generalization test. We present the average 
 best objective values and corresponding error bars over $51$ independent runs of all baselines in Table~\ref{tab:IID}. We additionally show the rank of each baseline on each tested problem instance and the averaged rank for intuitive performance comparison. From the results, we can observe that:

1) \textbf{Surr-RLDE v.s. Surr-RLDE-O.} Overall, Surr-RLDE surpasses optimization performance of Surr-RLDE-O, achieving second best average rank on the test set $P_{test}$~. This clearly demonstrate that at least in this case study, substituting the original function evaluation in the lower level of a MetaBBO method by a surrogate model of the target optimization problem does no harm to the overall mata-learning effectiveness. This is an appealing observation since as we elaborate in Section~\ref{sec3.2}, incorporating MetaBBO method with surrogate model in its low-level optimization could reduce the massive number of function evaluations needed to sample trajectory and train the meta-level policy. A particular interesting observation we would like to mention here is: the first four problems in Table~\ref{tab:IID} are multimodal problems with clear global structure, where our Surr-RLDE is only comparable with Surr-RLDE-O. However, when we look at the results for the last four problems which are multimodal problem with unclear global structure, Surr-RLDE show superior optimization performance. A possible explanation is that although the surrogate model show certain prediction error for the landscape of the target problem, this might be beneficial for problems with complex landscape. In this case our Kan-based surrogate serves as a simplification of the original landscape and provides  Surr-RLDE a shortcut to locate optima. Further studies on this point is needed and we mark it as an important future work.

2) \textbf{Surr-RLDE v.s. MetaBBO baselines.} GLEET as an up-to-date strong MetaBBO baseline achieves best performance against all baselines and our Surr-RLDE slightly underperforms it, with comparable even superior generalization performance on five of all eight tested problems. This further validate the effectiveness of the surrogate learning stage. By learning the landscape via the novel ROA loss we proposed, we ensure that the approximated landscape, especially the relative order information across different solutions is almost accurate and is enough to provide correct reward signal to assist the training of the meta-level policy. Our Surr-RLDE also outperforms DE-DDQN and DEDQN which employ similar value-based RL technique for dynamic algorithm configuration in DE algorithm. In particular, we observe that DE-DDQN achieve best performance on \emph{Schffers} and \emph{Schffershighcond} which are multimodal problems with global structure. Surr-RLDE achives best performance on \emph{Gallagher101Peaks} and \emph{Gallagher21Peaks} which are multimodal problems with weak global structure. This might indicate that Surr-RLDE shows certain advantages on solving more complex problems since De-DDQN's configuration space~(operator selection among several mutation operators) is much smaller than Surr-RLDE~(mutation configurations including both operator selection and parameter tuning). 


\subsubsection{Out-Of-Distribution Generalization}
For MetaBBO approach, a key performance metric to measure its usefulness is to measure the generalization of this approach towards those optimization problems which hold different optimization properties than the problem instances used for training~\cite{ma2024metabox}. This is so called out-of-distribution generalization aspect of a MetaBBO approach. In this experiment section, we compare such performance aspect of Surr-RLDE and other baselines. Specifically, we alternate properties of problems in the test set $P_{test}$ by adding random shift/rotation on decision variables or expanding its dimension from 10 to 30. We denote these two cases as ``10D-S/R'' and ``30D'' respectively. For MetaBBO baselines including Surr-RLDE, we directly use their models trained on the training set $P_{train}$ to the problems with modified properties. We only illustrate the optimization curves~(with error bars over 51 independent runs) of all baselines on four of the tested problems in Figure~\ref{fig:OOD} due to the space limitation. All results show that Surr-RLDE presents robust generalization performance on more complex and unseen problems. This further underscore the effectiveness of integrating well-trained surrogate model into low-level function evaluation process MetaBBO.

\subsection{Ablation Studies}
\subsubsection{Surrogate Model Choices}\label{sec:modelchoice}
\begin{figure}[t]
\centering
\subfigure[MSE accuracy]{
\label{fig-cec-g}
\includegraphics[width=0.42\linewidth]{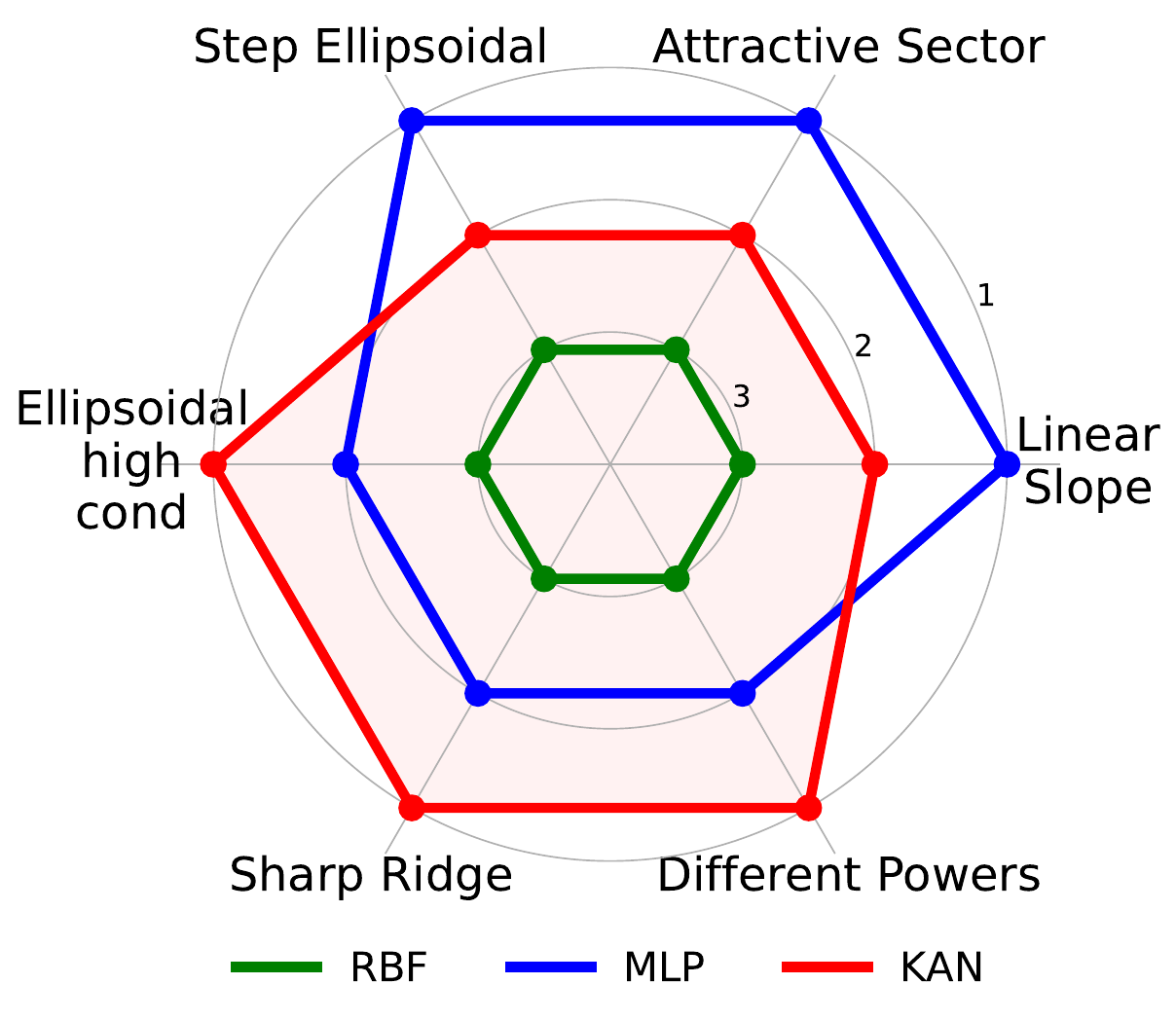}
}
\subfigure[ROA accuracy]{
\label{fig-hpo-g}
\includegraphics[width=0.42\linewidth]{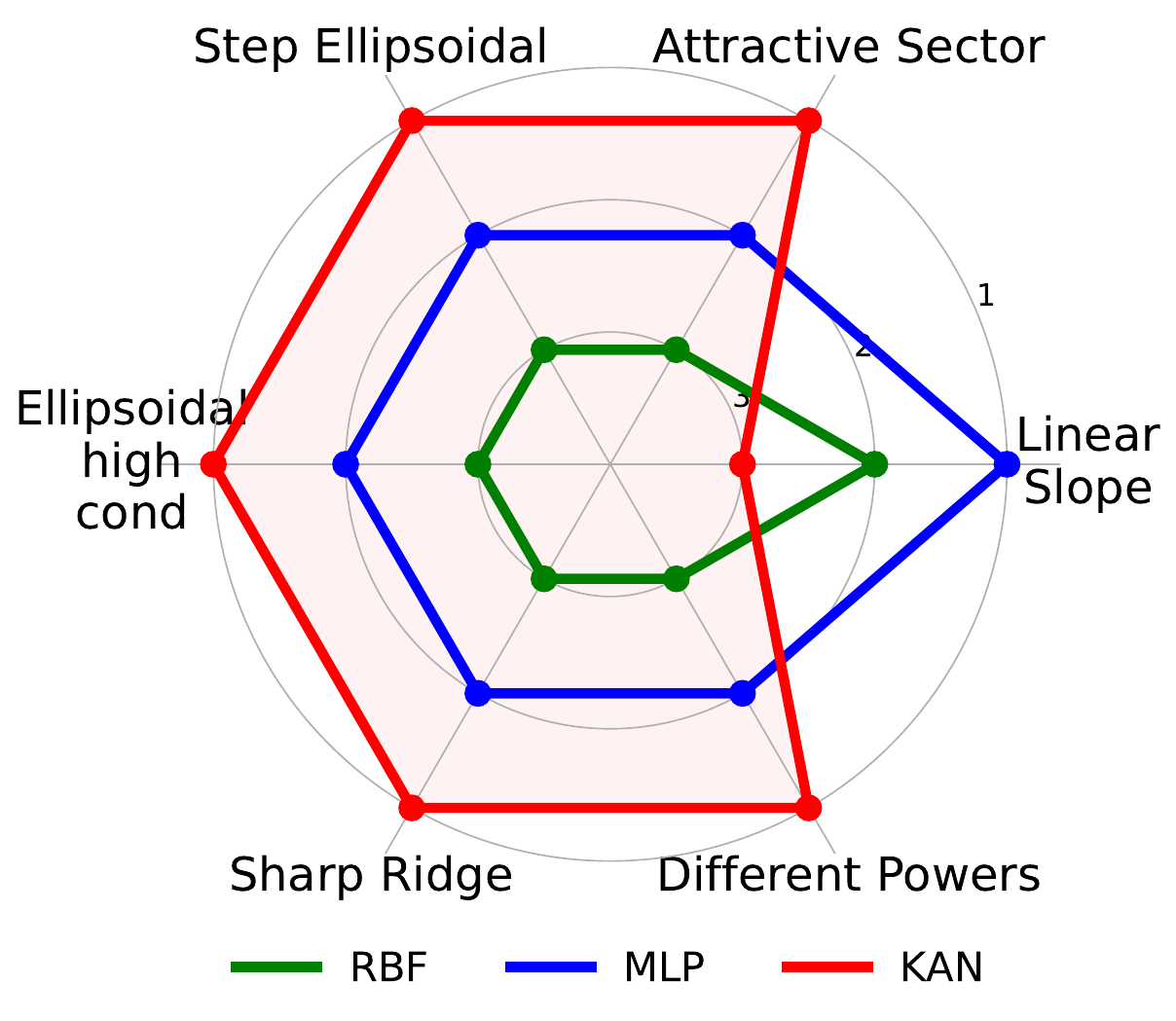}
}
\caption{Performance comparison of three ANN models.}
\vspace{-3mm}
\label{fig:ablation-network}
\end{figure}

We have to note that the selection of KAN as the surrogate model is non-trivial. In literature related with surrogate learning, we found that the dominating model choices stay in-between MLP and RBF networks, each of which excels in some applications. However, since MetaBBO is a novel task and KAN is an emerging network architecture, existing literature barely discuss these three models' performance differences and technical (dis-)~advantages. Hence we conduct a simple and intuitive comparison for these three models, where we train KAN, MLP and RBF as surrogate models for each problem instance in $P_{train}$ and illustrate the MSE loss rank and ROA loss rank of them on several problem instances in Figure~\ref{fig:ablation-network}. The results based on 200 independent repetitions show that: a) RBF network as a kernel-based~(e.g., Gaussian kernel) regression model could not achieve good prediction performance on the global landscape structure of complex optimization problem. b) In terms of MSE accuracy, MLP and KAN show similar prediction performance. c) However, KAN holds high-level ROA accuracy, which indicates that the spline-based univariate regression in KAN might be helpful to maintain local landscape, especially the relative order.       

\subsubsection{Relative-Order-Aware Loss}\label{sec:roaloss}
\begin{figure}[t]
\centering
\subfigure[\scriptsize{Rosenbrock\_original}]{
\includegraphics[width=0.3\linewidth]{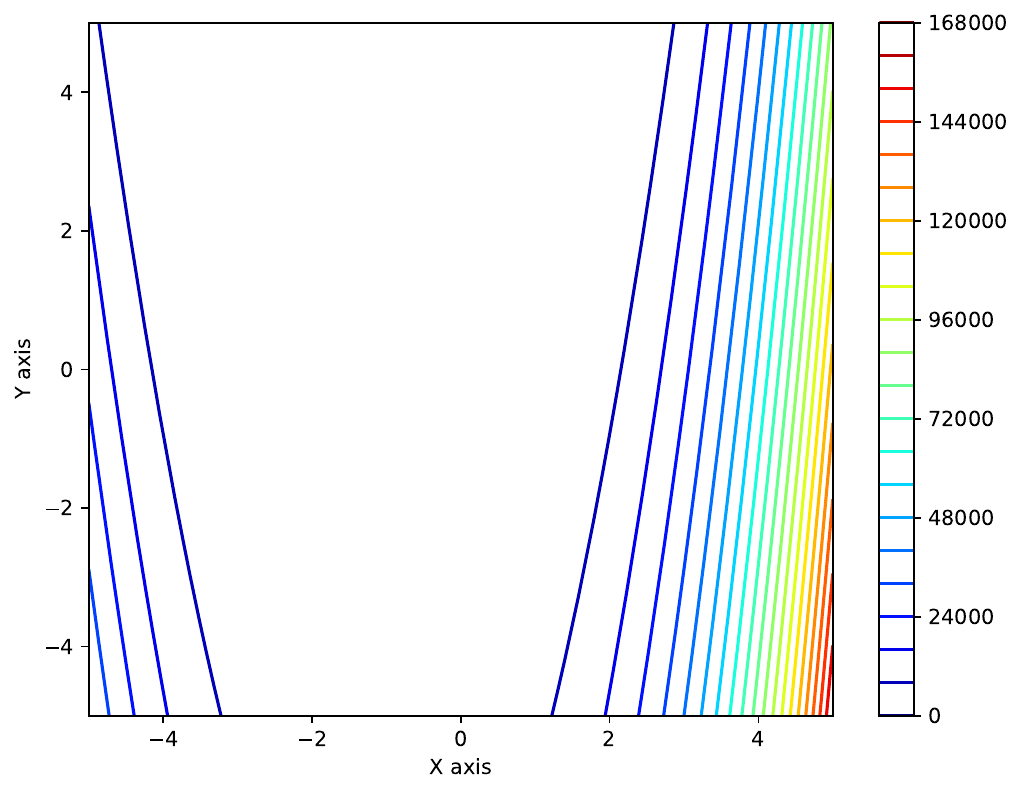}
}
\subfigure[\scriptsize{Fitted by MSE}]{
\includegraphics[width=0.3\linewidth]{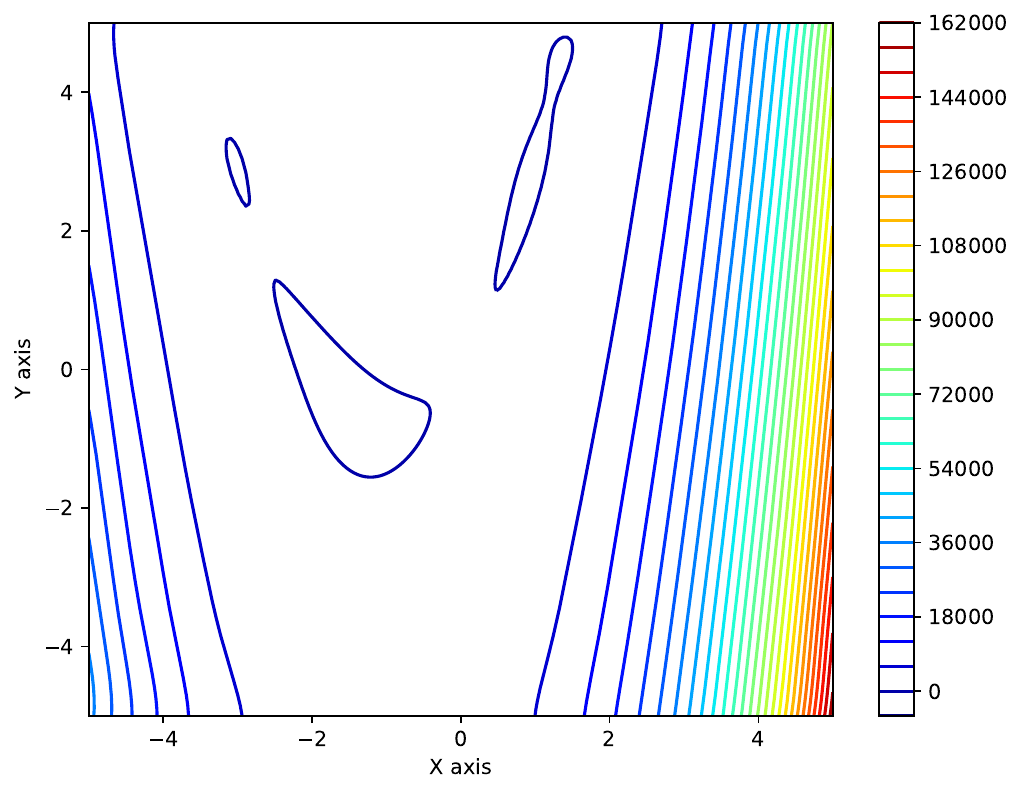}
}
\subfigure[\scriptsize{Fitted by ROA}]{
\label{fig-pd-g}
\includegraphics[width=0.3\linewidth]{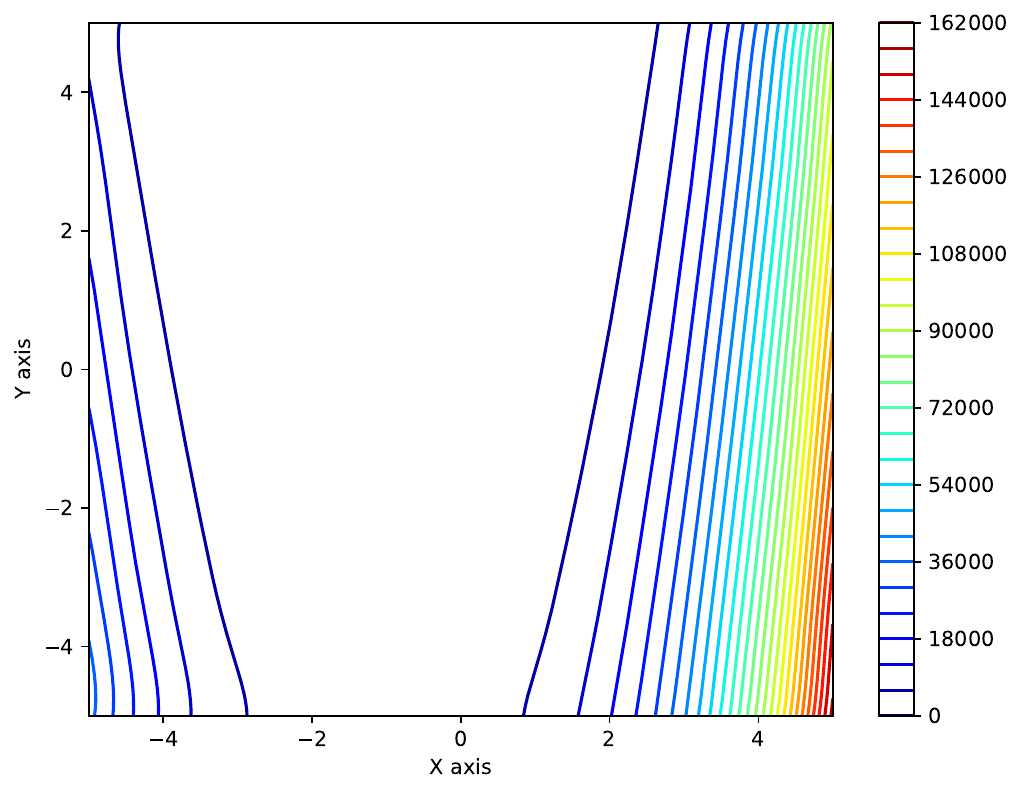}
}
\\
\subfigure[\scriptsize{Schwefel}]{
\label{fig-cec-s}
\includegraphics[width=0.3\linewidth]{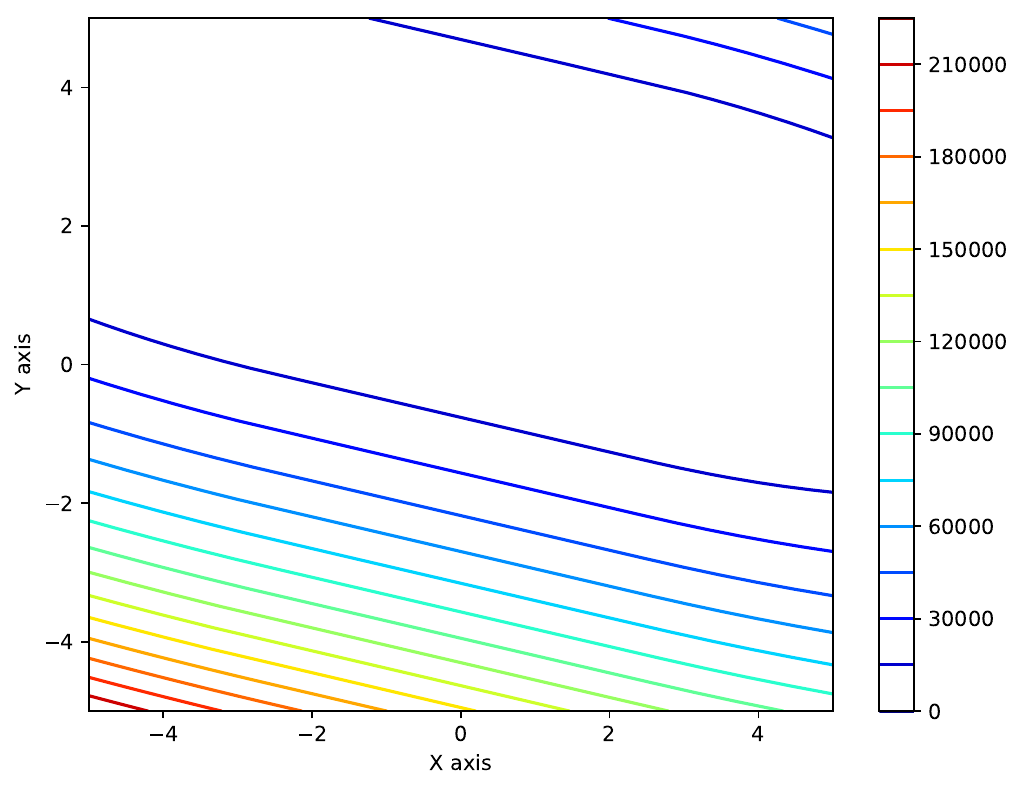}
}
\subfigure[\scriptsize{Fitted by MSE}]{
\label{fig-hpo-s}
\includegraphics[width=0.3\linewidth]{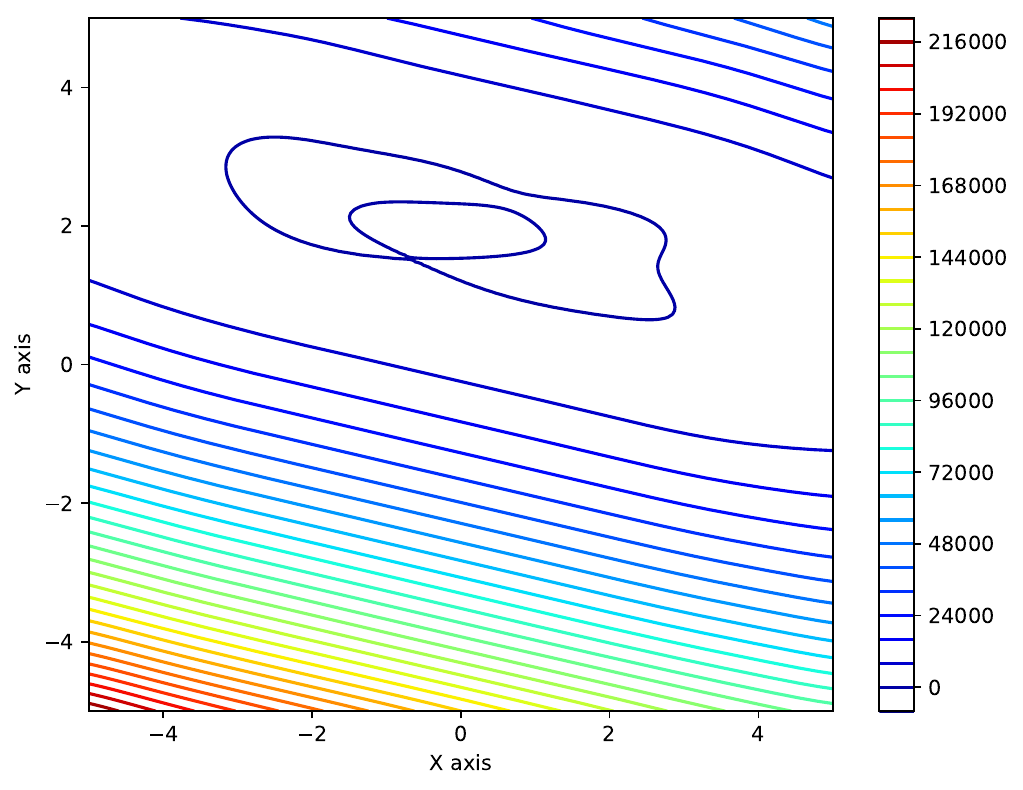}
}
\subfigure[\scriptsize{Fitted by ROA}]{
\label{fig-pd-s}
\includegraphics[width=0.3\linewidth]{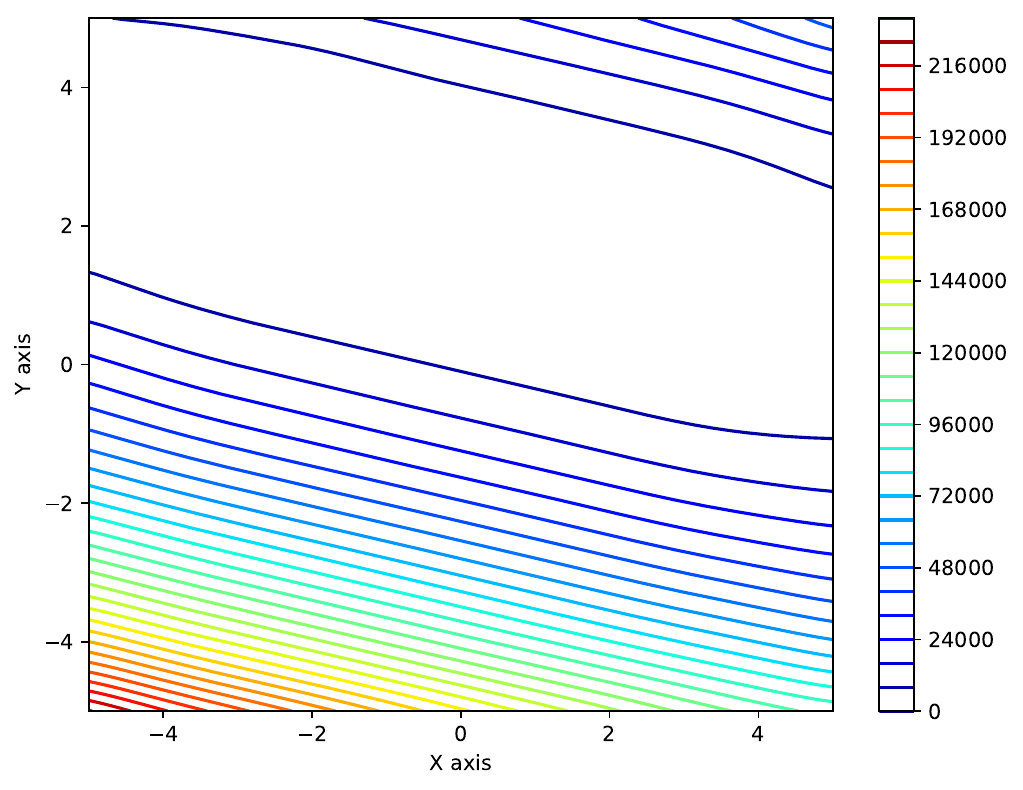}
}
\vspace{-1em}
\caption{Landscape accuracy of MSE and ROA loss.}
\vspace{-1em}
\label{fig:ablation-loss}
\end{figure}

Now we have determined the most suitable surrogate model design for Surr-RLDE: KAN architecture. We further conduct two studies on the effectiveness of the proposed ROA loss compared to naive MSE regression loss. We first visualize some qualitative evidence in Figure~\ref{fig:ablation-loss}, where we present two toy problems: \emph{Rosenbrock original} 2D and \emph{Schwefel} 2D as examples to show the landscapes predicted by the MSE-trained KAN model and the ROA-trained KAN model. The sample size for training KAN model on these two 2D problems is $10000$. We observe that the model trained by ROA loss could ensure the smoothness on the valley of two toy problems, which the model trained by MSE could not. This indicates that our proposed ROA loss, which adds a order correction term to the ordinary MSE loss, could improve the prediction accuracy by maintaining the local landscape structures. 
In contrast, MSE loss might only contribute to the global landscape fitting of the target problem. We further meta-train our Surr-RLDE and Surr-RLDE-MSE on $P_{train}$ and test the trained policies on the eight problems in $P_{test}$. The optimization results over $51$ runs are presented in Table~\ref{tab:ablation} and demonstrate that using surrogate models trained by ROA loss could provide more accurate reward signal than those trained by ordinary MSE loss.

\begin{table}[t]
\caption{Optimization performance comparison results of Surr-RLDE and Surr-RLDE-MSE on $P_{test}$}
\label{tab:ablation}
\resizebox{0.85\columnwidth}{!}{%
\begin{tabular}{c|c|c|c|c}
\hline
                               & \multirow{2}{*}{\textbf{Weierstrass}}         & \multirow{2}{*}{\textbf{Schaffers}}          & \multirow{2}{*}{\textbf{\begin{tabular}[c]{@{}c@{}}Schaffers\\ high cond\end{tabular}}} & \multirow{2}{*}{\textbf{\begin{tabular}[c]{@{}c@{}}Composite\\Grie rosen\end{tabular}}
                               } \\
\multicolumn{1}{l|}{}          &                                      &                                     &                                        &                                         \\ \hline
                               & mean                                 & mean                                & mean                                   & mean                                    \\
                               & std                                  & std                                 & std                                    & std                                     \\ \hline
\multirow{2}{*}{Surr-RLDE} & 6.914E+00                            & \cellcolor{gray!30}\textbf{4.330E-02}                  & \cellcolor{gray!30}\textbf{2.232E-01}                     & \cellcolor{gray!30}\textbf{1.769E+00}                      \\
                               & \small$(\pm$1.603E+00)  & \small($\pm$3.622E-02) & \small($\pm$1.007E-01)   & \small($\pm$3.189E-01)     \\ \hline
\multirow{2}{*}{Surr-RLDE-MSE} & \cellcolor{gray!30} \textbf{3.161E+00}                            & 2.936E+00                           & 8.653E+00                              & 1.791E+00                               \\
                               & \small($\pm$2.187E+00)  & \small($\pm$9.212E-01) & \small($\pm$3.756E+00)    & \small($\pm$9.466E-01)     \\ \hline
                               & \multirow{2}{*}{
                               \textbf{\begin{tabular}[c]{@{}c@{}}Gallagher\\101Peaks\end{tabular}}
                               } & \multirow{2}{*}{\textbf{\begin{tabular}[c]{@{}c@{}}Gallagher\\21Peaks\end{tabular}}
                               } & \multirow{2}{*}{\textbf{Katsuura}}              & \multirow{2}{*}{
                               \textbf{\begin{tabular}[c]{@{}c@{}}Lunacek bi\\Rastrigin\end{tabular}}
                               } \\
\multicolumn{1}{c|}{}          &                                      &                                     &                                        &                                         \\ \hline
\multirow{2}{*}{Surr-RLDE}                  & \cellcolor{gray!30}\textbf{3.448E-07}                   & \cellcolor{gray!30}\textbf{1.972E-05}                  & 1.458E+00                              & \cellcolor{gray!30}\textbf{4.245E+01}                      \\
                               & \small($\pm$1.071E-06)  & \small($\pm$1.277E-04) & \small($\pm$3.026E-01)    & \small($\pm$9.243E+00)     \\ \hline
\multirow{2}{*}{Surr-RLDE-MSE} & 5.234E+00                            & 1.217E+01                           & \cellcolor{gray!30}\textbf{5.576E-01}                              & 5.725E+01                               \\
                               & \small($\pm$3.316E+00)  & \small($\pm$1.148E+01) & \small($\pm$2.321E-01)    & \small($\pm$1.611E+01)     \\ \hline
\end{tabular}
}
\vspace{-1em}
\end{table}
\section{Conclusion}
In this paper, we conduct a preliminary study on feasibility of combining surrogate model learning with MetaBBO to achieve both effectiveness and efficiency. Specifically, we first identify the core limitation of ordinary MSE regression loss within surrogate learning: inability for maintaining accurate local landscape structure. To address this limitation, on the one hand, we adopt a novel neural network architecture KAN as the surrogate model which improves the surrogate's accuracy through spline-based univariate function composition. On the other hand, we design a relative-order-aware loss which forces the KAN model to learn the local landscape details. To verify if the proposed surrogate learning can assist MetaBBO, we design a simple MetaBBO task where the meta-level RL policy dynamically configures the low-level DE algorithm. Through comprehensive benchmarking, the feasibility of integrating the surrogate model into MetaBBO is sufficiently validated, including the contribution of both model selection and loss design. Surr-RLDE denotes a pivotal step towards MetaBBO with high efficiency, especially tailored for expensive, hard-to-evaluate BBO problems.

\begin{acks}
    This work was supported in part by the National Natural Science Foundation of China No. 62276100, in part by the Guangdong Provincial Natural Science Foundation for Outstanding Youth Team Project No. 2024B1515040010, in part by the Guangdong Natural Science Funds for Distinguished Young Scholars No. 2022B1515020049, and in part by the TCL Young Scholars Program.
\end{acks}


\bibliographystyle{ACM-Reference-Format}
\bibliography{myref}
\end{document}